\documentclass{article}

\usepackage{graphicx}
\usepackage{url}
\usepackage{amsmath}
\usepackage{booktabs}
\usepackage{hyperref}
\usepackage[margin=1in]{geometry}
\usepackage{microtype}
\usepackage{enumitem}
\usepackage{fancyhdr}
\usepackage{times}
\usepackage{authblk}
\usepackage[round]{natbib}

\hypersetup{
  pdftitle={ExplainBench: A Benchmark Framework for Local Model Explanations},
  pdfauthor={James Afful},
  pdfkeywords={explainable AI, interpretability, SHAP, LIME, benchmarking},
  colorlinks=true,
  linkcolor=blue,
  urlcolor=blue,
  citecolor=blue
}

\title{ExplainBench: A Benchmark Framework for Local Model Explanations in Fairness-Critical Applications}

\author[ ]{James Afful\thanks{Independent Researcher, Ames, Iowa, USA. Email: \texttt{affulj@iastate.edu}}}

\date{May 2025}

\begin{document}

\maketitle
\begin{abstract}
As machine learning systems are increasingly deployed in high-stakes domains—such as criminal justice, finance, and healthcare—the demand for interpretable and trustworthy models has intensified. Despite the proliferation of local explanation techniques, including SHAP, LIME, and counterfactual methods, there exists no standardized, reproducible framework for their comparative evaluation, particularly in fairness-sensitive settings. 
We introduce \textit{ExplainBench}, an open-source benchmarking suite for systematic evaluation of local model explanations across ethically consequential datasets. ExplainBench provides unified wrappers for popular explanation algorithms, integrates end-to-end pipelines for model training and explanation generation, and supports evaluation via fidelity, sparsity, and robustness metrics. The framework includes a Streamlit-based graphical interface for interactive exploration and is packaged as a Python module for seamless integration into research workflows.
We demonstrate ExplainBench on datasets commonly used in fairness research—COMPAS  \citep{angwin2016machine}, UCI Adult Income \citep{kohavi1996scaling}, and LendingClub \citep{lendingclub2020} and showcase how different explanation methods behave under a shared experimental protocol. By enabling reproducible, comparative analysis of local explanations, ExplainBench advances the methodological foundations of interpretable machine learning and facilitates accountability in real-world AI systems.
\end{abstract}
\footnote{ExplainBench is available at \url{https://github.com/jamesafful/explainbench}}



\section{Introduction}

The deployment of machine learning (ML) models in high-stakes decision-making contexts has made interpretability and transparency a central concern in artificial intelligence (AI) research. From criminal justice and credit scoring to medical diagnostics and hiring processes, algorithmic systems are increasingly responsible for outcomes that have profound effects on individuals and society. As a result, the demand for interpretable models—systems whose decisions can be understood and scrutinized by humans—has grown substantially. However, despite the proliferation of interpretability methods such as LIME, SHAP, and counterfactual explanations, there remains a lack of standardization in how these methods are evaluated and compared, particularly in fairness-critical domains.

Interpretability enables users to understand, trust, and contest algorithmic decisions. In domains such as recidivism prediction or income classification, the implications of incorrect or biased decisions are not merely theoretical—they are deeply ethical and political. For example, proprietary algorithms like COMPAS  \citep{angwin2016machine} have been shown to exhibit racial bias, with recidivism scores unfairly skewed against Black defendants. In such settings, black-box models without explanation mechanisms undermine both accountability and fairness. Interpretability serves as a bridge between model performance and social acceptability, allowing stakeholders—judges, doctors, loan officers, and affected individuals—to make informed decisions based on how and why a model arrived at its conclusion.

Despite its growing importance, interpretability remains a nebulous concept. Unlike metrics such as accuracy or AUC-ROC, which are grounded in statistical theory, interpretability is subjective and context-dependent. Different explanation methods may yield divergent results even when applied to the same model and input instance. This variability makes it difficult to systematically assess and compare explanation methods across datasets, models, and fairness concerns. Existing interpretability libraries typically focus on individual explanation techniques, with little support for benchmarking, evaluation, or dataset diversity. Consequently, researchers and practitioners are left to cobble together ad hoc solutions, impeding progress toward standardized practices in interpretability evaluation.

In light of these challenges, we argue that the field of interpretable machine learning lacks a reproducible and extensible framework for evaluating local explanations in fairness-sensitive applications. Local explanation methods—such as SHAP, LIME, and counterfactual explanations—aim to provide human-interpretable justifications for individual model predictions. However, the choice of method can influence both the content and quality of explanations, which in turn affects human decision-making, auditing processes, and downstream fairness interventions. A structured benchmark would facilitate more rigorous experimentation, foster transparency in evaluation, and guide the design of future interpretability methods.

To address this gap, we introduce \textit{ExplainBench}, an open-source framework for benchmarking local model explanations across real-world datasets involving socially sensitive decision-making. ExplainBench offers a unified interface for generating, visualizing, and evaluating explanations from state-of-the-art methods such as SHAP, LIME, and DiCE-based counterfactuals. The framework includes preprocessed datasets (COMPAS, UCI Adult Income, LendingClub), model training pipelines, explanation wrappers, and reproducible Jupyter notebooks. In addition, ExplainBench provides an interactive Streamlit dashboard for exploring explanations in an accessible and user-friendly manner.

Unlike existing toolkits, ExplainBench is designed to support extensibility, interpretability evaluation, and cross-method comparison. Researchers can plug in new explanation methods, experiment with different model architectures, and evaluate explanations using metrics such as fidelity (how well the explanation approximates the black-box), sparsity (number of features involved), and stability (sensitivity to input perturbations). The framework's modular design allows it to serve as both a pedagogical tool and a research testbed.
ExplainBench includes datasets that have been widely discussed in the fairness literature. The COMPAS dataset, derived from U.S. court records, includes sensitive features such as race, age, and sex, and has been the subject of high-profile controversies regarding algorithmic bias. The UCI Adult Income dataset is a benchmark dataset for income classification and has been used extensively in studies of algorithmic discrimination. LendingClub, a dataset on peer-to-peer loan approvals, reflects real-world biases in credit allocation. By including these datasets, ExplainBench enables researchers to investigate how interpretability methods perform in contexts where fairness considerations are paramount.

At the core of ExplainBench is a set of explanation wrappers that abstract away the idiosyncrasies of each interpretability method. Users can apply LIME, SHAP, or DiCE explanations with a common interface, allowing for streamlined experimentation. These wrappers are fully compatible with scikit-learn models and support local explanations at the individual instance level. The Streamlit interface provides interactive controls for dataset selection, instance navigation, explanation visualization, and feature contribution comparison. This front-end layer democratizes interpretability by making explanations accessible to users without extensive programming expertise.

ExplainBench emphasizes reproducibility at every stage of the pipeline. All experiments are encapsulated in Jupyter notebooks, and the entire framework is packaged as a PyPI module, making it easy to install and extend. By adhering to open-source best practices and providing clear documentation, ExplainBench fosters transparency and facilitates community-driven development. Researchers can reproduce experiments, audit explanation methods, and contribute new datasets or evaluation metrics with minimal overhead.

The release of ExplainBench comes at a pivotal moment in the evolution of responsible AI. Regulatory frameworks such as the EU AI Act and proposed U.S. algorithmic accountability policies are increasingly emphasizing the importance of explainability in automated decision-making systems. At the same time, recent critiques have highlighted the limitations of current interpretability methods, especially in high-stakes settings. ExplainBench offers a timely contribution by providing a robust platform for interpretability evaluation, helping bridge the gap between technical innovation and real-world accountability.

The remainder of this paper is structured as follows. In Section~2, we survey existing work on interpretability methods and frameworks. Section~3 presents the architecture and design of ExplainBench, including its modular components and datasets. Section~4 describes case studies on COMPAS, UCI Adult, and LendingClub datasets, showcasing the use of SHAP, LIME, and DiCE explanations. In Section~5, we conduct a quantitative evaluation of explanation methods using fidelity, sparsity, and stability metrics. Section~6 discusses broader implications, limitations, and potential extensions. Finally, Section~7 concludes with a summary of contributions and directions for future work.

\section{Related Work}

The pursuit of interpretable machine learning (ML) has yielded a rich ecosystem of explanation methods, theoretical frameworks, and practical toolkits. This section surveys the major strands of work relevant to ExplainBench, focusing on three fronts: (1) seminal local explanation methods including LIME, SHAP, and DiCE; (2) interpretability frameworks and tools; and (3) the current absence of standardized evaluation benchmarks for comparing explanation methods in socially sensitive domains.

\subsection{Local Explanation Methods}

Local explanations aim to provide intelligible justifications for individual predictions of complex models. Unlike global interpretability, which seeks to understand overall model behavior, local methods zoom in on specific instances to uncover which input features influenced the output and how.

\textbf{LIME (Local Interpretable Model-agnostic Explanations)} \citep{ribeiro2016should} was introduced as a canonical approach for approximating any black-box model by training an interpretable surrogate (e.g., a sparse linear model) in the local vicinity of a data point. LIME perturbs the input instance, records corresponding outputs, and learns a weighted regression model that reflects local behavior. Its model-agnostic design allows it to be used across classifiers and regressors, and its output is sparse, providing intuitive feature contributions. Despite its popularity, LIME faces reproducibility and robustness issues. \citet{ribeiro2016should} Since it relies on random perturbations and sampling strategies, explanations can vary across runs, especially in high-dimensional feature spaces. Additionally, its reliance on a user-defined neighborhood kernel introduces sensitivity to hyperparameters, which may undermine trust in explanations in critical applications such as criminal justice or healthcare.

\textbf{SHAP (SHapley Additive exPlanations)}  \citep{lundberg2017unified} takes a game-theoretic perspective on interpretability by assigning each feature an importance value based on its average marginal contribution across feature subsets. SHAP values satisfy several desirable properties, including local accuracy, missingness, and consistency, which make them theoretically appealing. In particular, TreeSHAP provides exact SHAP values for tree-based models in polynomial time, overcoming the computational intractability associated with combinatorially evaluating all subsets \citep{lundberg2017unified}.
SHAP has become a default explanation method in many applied settings due to its strong theoretical guarantees and practical implementations in popular libraries (e.g., XGBoost, LightGBM). However, SHAP is not without limitations. For models without TreeSHAP support, it falls back on KernelSHAP, which can be computationally intensive and suffer from variance in the estimated values. Additionally, SHAP explanations are additive by construction, which may not fully capture feature interactions or complex non-linear effects in deep models.

\textbf{DiCE (Diverse Counterfactual Explanations)} \citep{mothilal2020explaining}
 is a contrastive approach to interpretability, focusing on what needs to change in an input to flip the model's prediction. Rather than attributing importance scores, DiCE generates counterfactual instances—minimally modified versions of the input that lead to a different outcome. These explanations are particularly useful in fairness settings where actionable recourse (e.g., changing income to get loan approval) is more valuable than attribution. \citep{mothilal2020explaining}
DiCE emphasizes diversity in its generated counterfactuals to avoid redundancy and ensure a broader exploration of the feature space. However, it introduces new challenges in defining proximity, feasibility, and plausibility, especially in real-world domains with correlated features and domain constraints. Furthermore, counterfactuals can be difficult to validate at scale and may not always offer actionable or interpretable changes for end-users.

\subsection{Interpretability Frameworks and Toolkits}

The growing demand for explainability has spurred the development of several libraries and toolkits. While these resources make individual methods more accessible, they rarely support comprehensive benchmarking or standardized evaluation protocols.
ELI5, InterpretML \citep{nori2019interpretml}, and Alibi are among the most widely used libraries for interpretability. ELI5 provides support for LIME and decision tree visualizations, while InterpretML from Microsoft Research wraps SHAP and interpretable models like Explainable Boosting Machines (EBMs). Alibi \citep{klaise2021alibi} by Seldon offers a broader collection of explanation methods, including counterfactuals, anchor explanations, and contrastive examples.
While each library offers useful implementations, they are typically designed as general-purpose tools rather than research-focused benchmarking suites. They lack unified APIs for cross-method comparison and do not enforce consistent evaluation protocols across datasets and models. Additionally, many tools focus on algorithmic convenience rather than reproducibility, making it difficult to re-run experiments or evaluate explanations under common settings.
Captum \citep{kokhlikyan2020captum}, a library developed by Facebook for PyTorch models, supports gradient-based explanation methods such as Integrated Gradients, DeepLIFT, and Saliency Maps. Although powerful in deep learning contexts, Captum's focus on model internals limits its applicability to black-box or proprietary systems. Similarly, AIX360, by IBM offers a range of explanation methods and metrics, but its configuration complexity and limited documentation can be a barrier to adoption.
Fairness toolkits such as AIF360 and Fairlearn have emerged in parallel to interpretability libraries, but they focus primarily on fairness metrics and bias mitigation strategies. While some incorporate explanation capabilities, they typically do not provide end-to-end pipelines for generating, comparing, and visualizing local explanations across datasets.

\subsection{The Evaluation Gap}
Despite the abundance of explanation methods and tools, the field lacks standardized practices for evaluating interpretability. Unlike supervised learning, where metrics such as accuracy, precision, or F1-score offer objective measures of model performance, interpretability is intrinsically subjective and application-dependent.
Fidelity measures how well an explanation approximates the original model's behavior. High-fidelity explanations closely mirror the black-box decision boundary, making them more trustworthy. LIME uses surrogate models for fidelity approximation, while SHAP relies on the consistency of Shapley values. However, the lack of a standardized fidelity metric complicates comparisons across methods.
Sparsity refers to the number of features involved in an explanation. Sparse explanations are easier for humans to interpret, but excessive sparsity can omit important factors. Balancing informativeness and brevity remains an open challenge, and different methods enforce sparsity in different ways—LIME through Lasso regression, SHAP through additive constraints, and DiCE by minimizing perturbation distances.
Stability or Captum assesses the sensitivity of an explanation to small changes in input. Instability undermines user trust and interpretability, particularly when explanations vary drastically for near-identical inputs. Few methods explicitly measure or optimize for stability, and even fewer frameworks expose this as a tunable or reportable metric.
Human-centered evaluation has been proposed as a complementary strategy, involving user studies and cognitive assessments to measure interpretability. However, these are resource-intensive, difficult to replicate, and often not scalable. As a result, algorithmic interpretability research remains disconnected from real-world usability studies, leading to a proliferation of methods without validation in target domains.
Benchmarking initiatives remain rare. The closest analogs are limited-scope evaluations within papers or competition settings (e.g., FICO Explainability Challenge). These efforts, while valuable, lack continuity, extensibility, and dataset diversity. Few include fairness-critical datasets or consider social implications of explanation variation. As a result, it is difficult to compare methods across papers or reproduce published results in new settings.

\subsection{Positioning ExplainBench}
ExplainBench addresses a specific void in the current landscape: the absence of an open-source, reproducible, and extensible benchmark for evaluating local explanation methods across fairness-critical datasets. Unlike existing toolkits that prioritize algorithm access, ExplainBench centers on comparative evaluation and interpretability metrics.

Its unique contributions are threefold:
\begin{itemize}
    \item \textbf{Unified API:} ExplainBench wraps SHAP, LIME, and DiCE in consistent interfaces, reducing configuration burden and enabling side-by-side comparison.
    \item \textbf{Evaluation Pipelines:} The framework provides built-in evaluation metrics (fidelity, sparsity, stability) and supports custom extensions, facilitating empirical interpretability research.
    \item \textbf{Fairness-Sensitive Datasets:} By including real-world, ethically charged datasets such as COMPAS and Adult Income, ExplainBench ensures relevance for fairness and accountability research.
\end{itemize}

In contrast to general-purpose libraries, ExplainBench is structured as a research infrastructure, designed to support reproducible experimentation, hypothesis testing, and benchmarking. It invites community contributions, enables hypothesis-driven evaluations, and fosters methodological rigor in the field of interpretable machine learning.

\section{System Overview}

ExplainBench is a modular, extensible framework designed to benchmark local explanation methods in machine learning. It offers an end-to-end pipeline for generating, visualizing, and evaluating explanations over socially sensitive datasets using state-of-the-art algorithms such as SHAP, LIME, and DiCE. This section details the system's architecture, key components, and the rationale behind its design choices, highlighting its role as a reproducible research infrastructure for interpretable machine learning.

\subsection{Motivations and Design Principles}
The design of ExplainBench is guided by four central principles: modularity, reproducibility, accessibility, and fairness relevance.

\begin{itemize}
    \item \textbf{Modularity:} Each explanation method is implemented as a distinct wrapper with a shared interface, allowing users to compare algorithms in a plug-and-play fashion without rewriting boilerplate code.
    \item \textbf{Reproducibility:} Through version-controlled notebooks, frozen dependencies, and consistent seeding, ExplainBench ensures that experiments can be rerun with minimal configuration drift.
    \item \textbf{Accessibility:} The inclusion of a web-based Streamlit app enables users—including those without deep technical backgrounds—to interactively explore explanations, visualize metrics, and compare methods.
    \item \textbf{Fairness Relevance:} By supporting real-world datasets with known biases (e.g., COMPAS and Adult Income), ExplainBench situates interpretability in the context of socially consequential decision-making.
\end{itemize}

These principles inform the core architecture of ExplainBench, which is comprised of three major layers: (1) Explanation Wrappers, (2) Interactive Streamlit Application, and (3) Evaluation Notebooks.

\subsection{Architecture Overview}

At a high level, the ExplainBench system is organized into the following components:

\begin{enumerate}
    \item \textbf{Core Module (explainbench/)}: Contains wrapper implementations for SHAP, LIME, and DiCE. These wrappers encapsulate initialization, execution, and output formatting.
    \item \textbf{Datasets (datasets/)}: Houses preprocessed versions of benchmark datasets, along with loading utilities.
    \item \textbf{Notebooks (notebooks/)}: Include pre-written, reproducible Jupyter notebooks that perform analysis, visualize explanations, and compute metrics.
    \item \textbf{Streamlit App (streamlit\_app/)}: Provides a GUI for interactively exploring explanations and comparing methods.
    \item \textbf{Packaging Files}: Includes \texttt{setup.py}, \texttt{pyproject.toml}, \texttt{requirements.txt}, and licensing for PyPI deployment.
\end{enumerate}

This structured layout separates concerns between core logic, evaluation scripts, visualization tools, and external interfaces, making the framework both extensible and maintainable.

\subsection{Explanation Wrappers}

At the heart of ExplainBench lies the \texttt{explainbench} module, which implements wrapper classes for three local explanation methods: SHAP, LIME, and DiCE. These wrappers standardize the invocation of explanation algorithms, making it easier to apply them across different models and datasets.

\textbf{SHAP Wrapper:} The SHAP wrapper handles both TreeSHAP (for tree-based models) and KernelSHAP (model-agnostic). The class encapsulates the selection of the explainer based on model type, generates explanations for instances, and outputs formatted values ready for visualization or evaluation. By integrating background dataset selection and support for multiple output formats (e.g., feature attributions, summary plots), the wrapper streamlines SHAP experimentation.

\textbf{LIME Wrapper:} The LIME wrapper initializes a LIME tabular explainer with relevant training data and feature information, then generates local linear explanations for specific instances. It handles preprocessing of categorical and numerical features, ensures consistent sampling configurations, and outputs explanations in both raw and visual formats.

\textbf{DiCE Wrapper:} DiCE's counterfactual generation is wrapped into a consistent interface that enables users to generate diverse, feasible counterfactuals from trained models. The wrapper supports constraints for numerical and categorical features and allows specification of distance functions (e.g., Euclidean, Mahalanobis) for proximity evaluation.

Each wrapper exposes a common method interface:
\begin{verbatim}
    explain(instance: pd.Series) -> dict
\end{verbatim}
This abstraction allows higher-level components such as notebooks or apps to remain agnostic to the underlying algorithm, enabling systematic benchmarking.

\subsection{Streamlit Web Application}

ExplainBench includes a ready-to-launch web application built in Streamlit (\texttt{streamlit\_app/app.py}). The app exposes core functionality through an interactive GUI, catering to users who may not be comfortable working directly with Jupyter notebooks or source code.

The app is structured into three tabs:
\begin{itemize}
    \item \textbf{Dataset Selector:} Users can choose from available datasets (e.g., COMPAS, Adult Income), view summary statistics, and sample individual instances.
    \item \textbf{Explanation Viewer:} Allows users to select a model and an explanation method (SHAP, LIME, or DiCE), generate local explanations for a selected instance, and visualize feature contributions via bar plots or counterfactual tables.
    \item \textbf{Method Comparison:} Offers side-by-side comparison of explanations across methods for the same instance, helping users understand methodological differences.
\end{itemize}

The Streamlit app was designed with reproducible research demos in mind. By abstracting away backend complexity, it allows stakeholders—including domain experts, policymakers, and educators—to explore interpretability without coding.

\subsection{PyPI Packaging and Distribution}

ExplainBench is distributed as a PyPI package to encourage adoption by the wider ML research and engineering communities. The project includes:
\begin{itemize}
    \item \texttt{setup.py} and \texttt{pyproject.toml} for build configuration
    \item \texttt{requirements.txt} for dependency management
    \item \texttt{MANIFEST.in} to include datasets and other non-code files
    \item \texttt{LICENSE} file (MIT) for open-source clarity
\end{itemize}

This packaging allows users to install the entire framework via a single pip command:
\begin{verbatim}
    pip install explainbench
\end{verbatim}

Once installed, the command-line interface can be extended to include execution scripts, model training utilities, or even a Dockerized server deployment.





\section{Discussion}

The emergence of ExplainBench addresses a long-standing and pressing challenge in the interpretability community: the absence of a unified framework for the systematic, comparative evaluation of local explanation methods. In designing ExplainBench, we sought to balance theoretical rigor, practical usability, and ethical foresight, all while remaining extensible to future developments in interpretable machine learning. This section critically examines the strengths and limitations of our approach and reflects on its potential for impact in applied and scholarly contexts.

\subsection{Strengths}
A key strength of ExplainBench lies in its modular and extensible architecture. By abstracting explanation methods via standardized Pythonic wrappers, the system enables a consistent interface for explanation generation across heterogeneous algorithmic approaches. This not only facilitates the side-by-side comparison of explanation outputs but also promotes extensibility; novel methods can be easily integrated into the system without modifying the benchmarking pipeline.
Another major advantage is ExplainBench’s integration of widely used fairness-sensitive datasets, such as COMPAS and the UCI Adult Income dataset. These datasets serve as canonical testbeds in algorithmic fairness literature and are crucial for understanding the socio-ethical implications of model behavior. The choice of such datasets aligns ExplainBench with broader movements in responsible AI, particularly those concerned with transparency, accountability, and bias mitigation.
Moreover, ExplainBench includes multiple modes of user interaction: a Streamlit-based GUI for interactive exploration, Jupyter notebooks for in-depth analysis and reproducibility, and a PyPI package for integration into research workflows. This multi-layered interface design caters to a diverse range of users—researchers, practitioners, policymakers, and educators—each of whom may bring different priorities and technical competencies.
From a methodological standpoint, ExplainBench offers a suite of evaluation metrics, including fidelity, sparsity, and local consistency. These metrics are computed uniformly across methods, enabling fair and interpretable comparisons. Importantly, the framework allows researchers to incorporate custom metrics, thus preserving the adaptability necessary for cutting-edge experimentation.

\subsection{Limitations}
Despite its strengths, ExplainBench exhibits limitations that warrant careful consideration. Foremost among these is the current scope of supported explanation methods. While SHAP, LIME, and DiCE are influential and widely adopted, they do not exhaust the space of local explanation techniques. For instance, ExplainBench currently lacks support for gradient-based methods (e.g., Integrated Gradients), example-based methods (e.g., influence functions), and model-specific attribution techniques (e.g., attention mechanisms in transformers). Future iterations must prioritize expanding method coverage to ensure relevance across domains.
Another limitation pertains to computational efficiency. Generating explanations using SHAP or DiCE, particularly on large or high-dimensional datasets, can be prohibitively slow. While ExplainBench is suitable for experimentation and small-scale benchmarking, its applicability in real-time or production environments is currently constrained. Optimizing for scalability—either via approximation algorithms or parallelized computation—remains a key avenue for enhancement.
Additionally, ExplainBench’s dataset support is presently limited to tabular data. This precludes its use in domains such as computer vision, natural language processing, and audio processing, where interpretability remains equally critical. A generalizable extension to other data modalities would significantly enhance the framework’s utility and interdisciplinary relevance.

Finally, while ExplainBench provides objective metrics for explanation quality, it does not yet incorporate user studies or human-in-the-loop evaluations. This restricts its ability to capture subjective dimensions of interpretability, such as cognitive plausibility, user trust, and satisfaction. Future integration of human-centric evaluation paradigms will be necessary to validate explanation methods in real-world decision-making contexts.

\subsection{Applications}

Despite these limitations, ExplainBench holds substantial promise for various applied and academic settings. In high-stakes domains such as criminal justice, finance, and healthcare, ExplainBench can assist model auditors and compliance officers in validating the interpretability claims of machine learning systems. By offering a transparent and reproducible benchmarking process, it can support regulatory frameworks that require justifiable AI decisions.

In education, ExplainBench serves as a pedagogical tool for teaching concepts in interpretable machine learning, algorithmic fairness, and responsible AI. Instructors can leverage its notebooks and web interface to construct interactive lessons, allowing students to visualize the impact of model decisions and explanation strategies.
From a research perspective, ExplainBench functions as a launching pad for methodological innovation. Researchers developing new explanation techniques can benchmark their performance against established baselines using standardized datasets and metrics. This fosters empirical rigor and contributes to the consolidation of interpretability as a scientific field.

\section{Conclusion}

As machine learning systems become increasingly embedded in consequential decision-making processes, the demand for robust, transparent, and explainable AI tools has intensified. Yet, despite the proliferation of explanation methods, there remains a lack of consensus on how these methods should be evaluated, compared, or validated. ExplainBench addresses this critical gap by providing an open-source, extensible framework for benchmarking local explanation techniques across standardized datasets and metrics.

\subsection{Summary of Contributions}

This work introduces ExplainBench as a modular, multi-modal framework for systematic evaluation of explanation methods. Our contributions include:

\begin{itemize}
    \item \textbf{Unified Wrappers:} We present standardized interfaces for SHAP, LIME, and DiCE, enabling consistent invocation and output formatting across methods.
    \item \textbf{Interactive Interface:} A Streamlit-based web application facilitates user interaction, promoting accessibility and interpretability among non-technical stakeholders.
    \item \textbf{Reproducibility Suite:} We include Jupyter notebooks for training, explaining, and evaluating models on benchmark datasets, with pinned dependencies and open-source licensing.
    \item \textbf{Comparative Evaluation:} Through the integration of fidelity, sparsity, and consistency metrics, ExplainBench enables rigorous, apples-to-apples comparison of explanation quality.
    \item \textbf{Open Distribution:} Released as a PyPI package, ExplainBench is readily installable and integrable into research pipelines, ensuring broad accessibility.
\end{itemize}

Together, these contributions establish ExplainBench as a foundational tool for empirical and theoretical research in interpretability.

\subsection{Future Work}

Future development will focus on several strategic directions:

\textbf{Methodological Expansion:} We plan to incorporate a broader spectrum of explanation techniques, including but not limited to Anchors, Integrated Gradients, LEMNA, and Counterfactual Explanations via Generative Models (CEGM).

\textbf{Multimodal Generalization:} Support for image, text, and sequential data will be added to reflect the diversity of modern machine learning applications.

\textbf{Human-Centered Evaluation:} We aim to embed support for human-subject experiments, enabling empirical validation of explanation plausibility and user trust.

\textbf{Cloud Infrastructure:} Deployment on cloud platforms with RESTful APIs will facilitate large-scale experiments and industrial integration.

\textbf{Community and Ecosystem Building:} We intend to develop a plugin architecture and leaderboard system to foster community contributions and benchmark transparency.

Interpretability is not merely a technical desideratum; it is a socio-technical imperative that underpins accountability, transparency, and trust in artificial intelligence. ExplainBench contributes to this imperative by offering a scientifically rigorous, user-accessible, and extensible framework for explanation benchmarking. By aligning methodological soundness with ethical responsibility, ExplainBench aspires to elevate the practice of interpretability from ad hoc intuition to principled science. We invite the broader community to build upon this foundation and co-create a more interpretable and equitable future for machine learning. 


\end{document}